\pdfoutput=1
\documentclass[10pt,twocolumn,letterpaper]{article}

\usepackage{iccv}
\usepackage{times}
\usepackage{epsfig}
\usepackage{graphicx}
\usepackage{amsmath}
\usepackage{amssymb}  

\usepackage{mathtools}
\usepackage{bm}
\usepackage{esvect}

\usepackage{algorithm}
\usepackage{algorithmicx}
\usepackage{algpseudocode}
\usepackage{mathtools}
\usepackage{url}
\usepackage[table]{xcolor} 
\usepackage[para,online,flushleft]{threeparttable}
\usepackage{arydshln}
\usepackage[pagebackref=true,breaklinks=true,colorlinks,bookmarks=false]{hyperref}
\usepackage{caption}
\usepackage{subcaption}
\usepackage{makecell}
\usepackage{pifont}
\usepackage{multirow}
\usepackage{booktabs}
\usepackage{chngcntr}
\usepackage{authblk}

\newcommand\blfootnote[1]{%
  \begingroup
  \renewcommand\thefootnote{}\footnote{#1}%
  \addtocounter{footnote}{-1}%
  \endgroup
}

\iccvfinalcopy 
\def\iccvPaperID{1663} 

\ificcvfinal\fi

\begin{document}

\title{TrackMPNN: A Message Passing Graph Neural Architecture for \\Multi-Object Tracking}

\author[1]{Akshay~Rangesh}
\author[2]{Pranav~Maheshwari}
\author[2]{Mez~Gebre}
\author[2]{Siddhesh~Mhatre}
\author[2]{Vahid~Ramezani}
\author[1]{Mohan~M.~Trivedi}
\affil[ ]{\textit{\{arangesh, mtrived\}@ucsd.edu, \{pranav, mez, siddhesh, vahid.ramezani\}@luminartech.com}}
\affil[1]{Laboratory for Intelligent \& Safe Automobiles, UC San Diego, CA}
\affil[2]{Luminar Technologies, Inc., Palo Alto, CA}

\maketitle

\blfootnote{$^\ddagger$Code: \href{https://github.com/arangesh/TrackMPNN}{https://github.com/arangesh/TrackMPNN}}
\ificcvfinal\thispagestyle{empty}\fi

\begin{abstract}
This study follows many classical approaches to multi-object tracking (MOT) that model the problem using dynamic graphical data structures, and adapts this formulation to make it amenable to modern neural networks. Our main contributions in this work are the creation of a framework based on dynamic undirected graphs that represent the data association problem over multiple timesteps, and a message passing graph neural network (MPNN) that operates on these graphs to produce the desired likelihood for every association therein. We also provide solutions and propositions for the computational problems that need to be addressed to create a memory-efficient, real-time, online algorithm that can reason over multiple timesteps, correct previous mistakes, update beliefs, and handle missed/false detections. 
To demonstrate the efficacy of our approach, we only use the 2D box location and object category ID to construct the descriptor for each object instance. Despite this, our model performs on par with state-of-the-art approaches that make use of additional sensors, as well as multiple hand-crafted and/or learned features. This illustrates that given the right problem formulation and model design, raw bounding boxes (and their kinematics) from any off-the-shelf detector are sufficient to achieve competitive tracking results on challenging MOT benchmarks.
\end{abstract}

\section{Introduction}\label{sec:introduction}

%
%
%
%
Object tracking is a crucial part of many complex autonomous systems and finds application in a variety of domains that require tracking of humans and/or objects. Tracking introduces memory and persistence into a system via association, as opposed to standalone measurements and detections in time. This enables systems to monitor object or agent behavior over a period of time, thus capturing historical context which can then be used to better assess current state or more accurately predict future states. Examples of such usage from different domains include - gauging human/pedestrian intent and attention based on history of movement~\cite{deo2020trajectory}\cite{ohn2016looking}\cite{ridel2018literature}, surveillance of crowds scenes to perceive crowd behavior~\cite{swathi2017crowd}, predicting future trajectories of vehicles based on past tracks~\cite{sivaraman2013looking}\cite{deo2018would}, tracking compact entities like cells and molecules to better understand biological processes~\cite{he2017cell} etc. 

\begin{figure}[t]
\begin{center}
\includegraphics[width=0.75\linewidth]{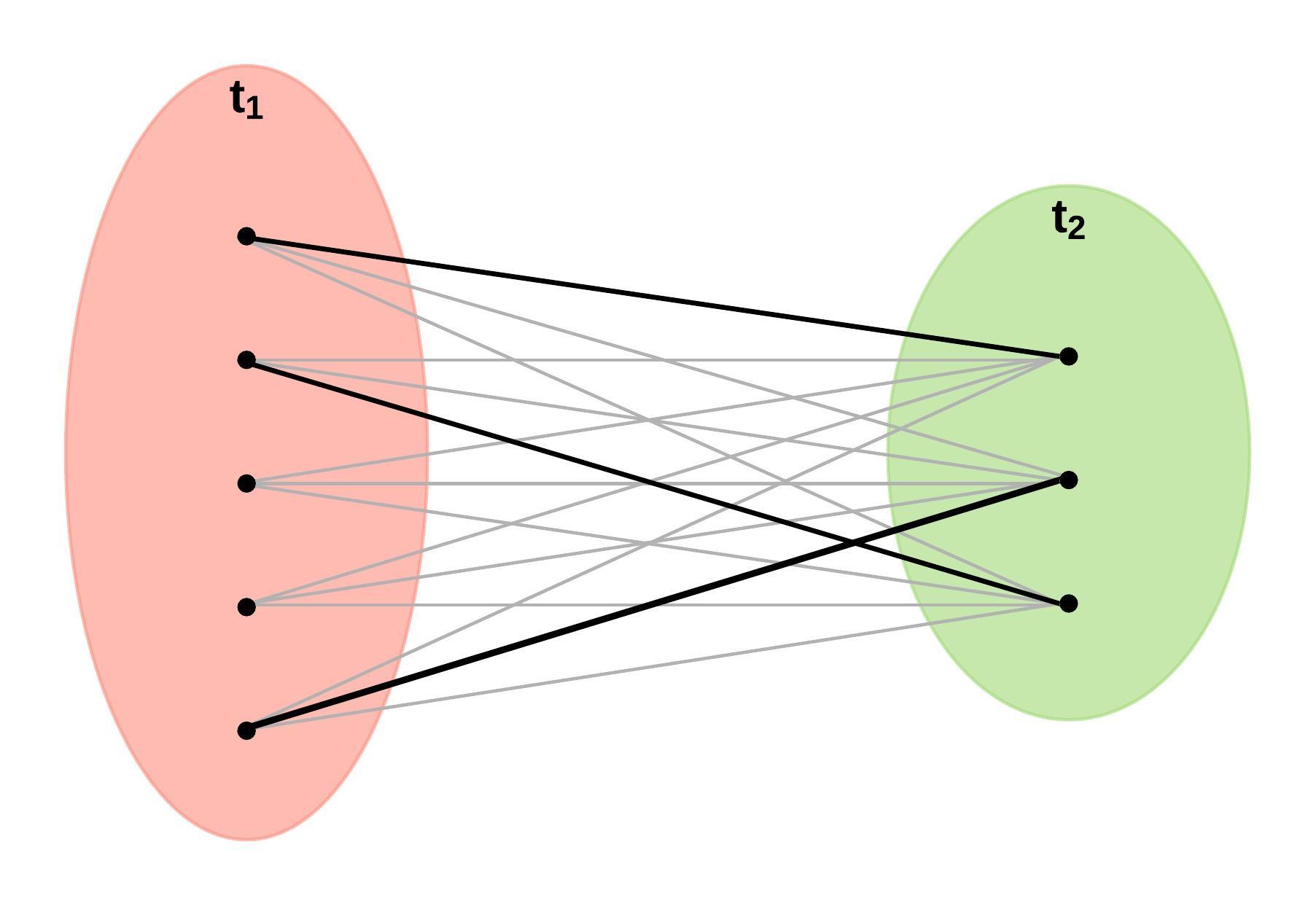}
\end{center}
\caption{Most tracking-by-detection approaches are reduced to successive bipartite matching problems between two disjoint sets, namely - sets of detections from two (consecutive) timesteps as shown in the illustration above. The goal of the multi-object tracker is to propose correct associations between the two sets (bold lines), and suppress spurious ones (gray lines).}
\label{fig:bipartite}
\end{figure}

Tracking methods are numerous and varied in their approach, and can broadly be categorized based on the number of objects being tracked i.e. single or multi-object tracking (MOT) approaches. MOT approaches go beyond feature and appearance tracking and attempt to solve successive data association problems to resolve conflicts between multiple competing observations and tracks. Thus, the problem of MOT is also a problem of data association - where multiple observations are to be assigned to multiple active object tracks to optimize some global criterion of choice. MOT approaches can further be classified based on their mode of operation i.e. online, offline (batch), or near-online. Online approaches provide tracks in real-time and continuously, and thus find use in real-time systems like autonomous cars. On the other hand, offline approaches benefit from viewing an entire sequence (i.e. conditioning on all available data) before estimating object tracks. This makes them relatively more robust than their online counterparts. However, these approaches are only suited to applications where post hoc operation is not a hindrance, e.g. surveillance.

Modern MOT techniques typically work within the tracking-by-detection paradigm, where trackers function by stitching together individual detections across time (see Figure~\ref{fig:bipartite}). These detections can either be obtained from a separate upstream object detector that operates independently, or by integrating the detector and tracker into one cohesive unit. Irrespective of the particularities of the approach, multi-object trackers aim to robustly track multiple objects through appearance changes, missed/false detections, temporary occlusions, birth of new tracks, death of existing tracks, and other such inhibitors. They do this by engineering better feature descriptors for each object, improving the similarity or cost function used during data association, or by better optimizing the underlying objective.

In this study, we propose a novel MOT framework based on dynamic, undirected, bipartite graphs for contemporary neural networks. Our approach is partly motivated by the probabilistic graph representation of classical multi-target tracking methods like Multi-Hypothesis Tracking (MHT)~\cite{cox1996}, with a focus on compatibility with modern graph neural networks (GNNs). Our approach is online, capable of tracking multiple objects, reasoning over multiple timesteps, and holding multiple hypotheses at any given time. In addition to this, our proposed approach can run real-time on almost any modern GPU without exceeding memory and compute limits.

\begin{figure*}[t]
    \center{\includegraphics[width=0.85\textwidth]
    {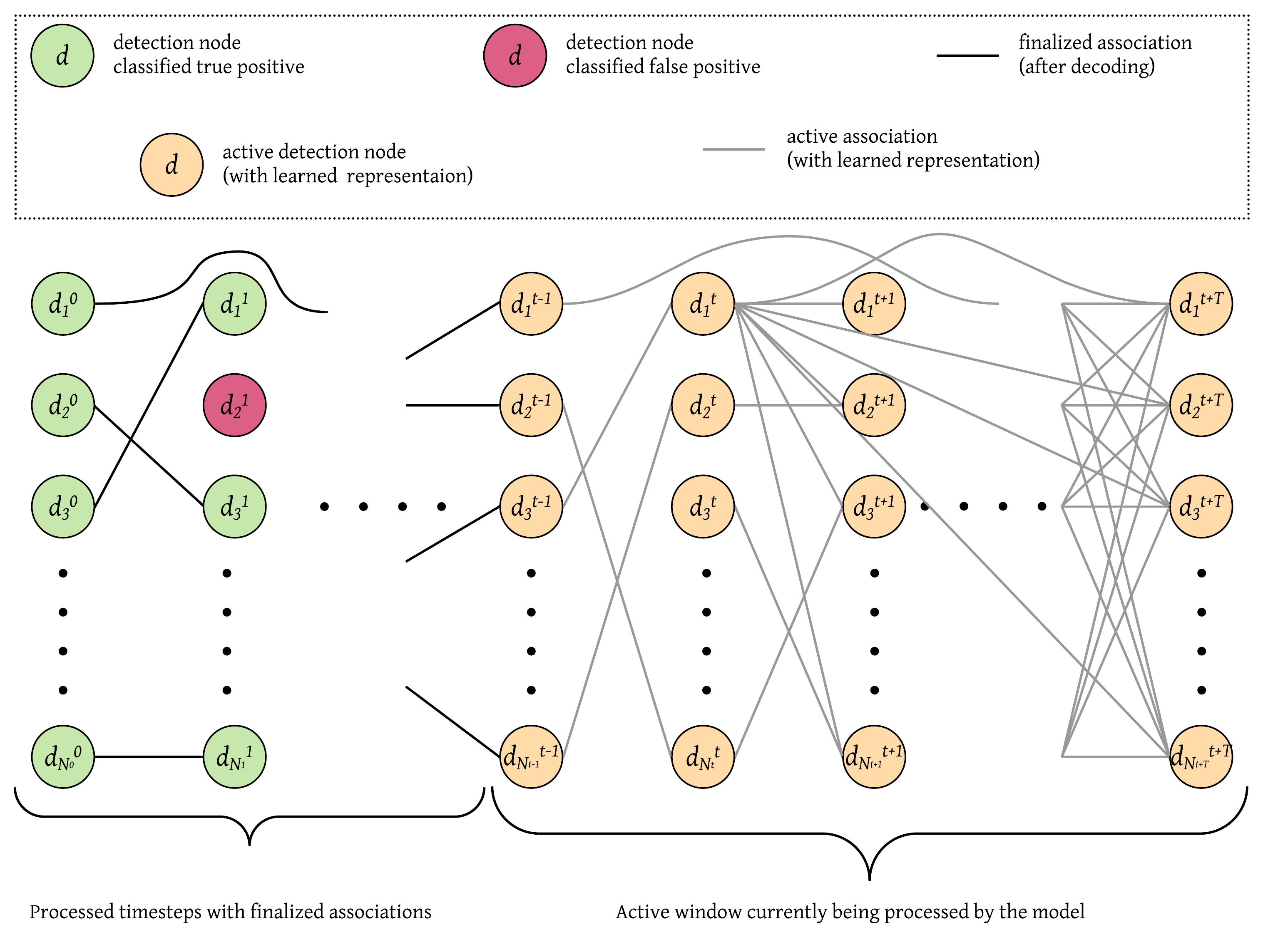}}
    \caption{\label{fig:overview} Our proposed MOT framework is based on dynamic undirected graphs that depict the data association problem over multiple timesteps. In the Figure above, $d_i^j$ represents a detection node indexed by $i$, at timestep $j$.}
\end{figure*}

\section{Related Research}

Modern MOT techniques have different approaches to address the underlying problem of data association. While most methods utilize some form of tracking-by-detection, some create a cohesive framework where the detector and tracker work in tandem~\cite{zhang2020fairmot}\cite{Hu3DT19}\cite{zhou2020tracking}. This usually leads to a symbiotic relationship, where detector and tracker benefit one another. In similar fashion, some studies have explored the benefits of incorporating other related tasks in their MOT framework, including segmentation~\cite{zeng2019dmm}\cite{Voigtlaender2019CVPR} and reconstruction\cite{luiten2019MOTSFusion}. Some others tweak classical approaches by making them more tractable~\cite{hamid2015joint}, or by developing better feature representations~\cite{kim2015multiple} - thereby making them amenable for deployment in dense, cluttered scenes. Other works focus solely on feature engineering and tuning of cost functions~\cite{MOTBeyondPixels,karunasekera2019multiple}, and assume detections from a trained off-the-shelf detector. Finally, with the introduction of large scale datasets for autonomous driving, many modern approaches tend to make use of 3D sensors like LiDARs, either by itself~\cite{poschmann2020factor}, or via multi-modal fusion with other imaging sensors like cameras~\cite{rangesh2019no}. This helps these methods resolve temporary occlusion, truncation, and other such issues arising from 3D ambiguity. 

In this study, we introduce a MOT framework suitable for Graph Neural Networks (GNNs). GNNs have found a wide range of applications in recent years; see \cite{zhou2019} and references therein. GNNs can be viewed as a generalization of convolutional neural networks(CNNs) \cite{shea2015}. Multi-layer CNNs can extract multi-scale  features from spatio-temporal data and generate  representations which can solve a variety of machine learning tasks. However, CNNs are defined by kernels that operate only on regular multi-dimensional grids. GNNs generalize CNNs kernel operation to general graphs (of which CNNs on regular grids are special cases). More importantly for us, one can interpret the kernel operations in this graphical context as generating an output at each node using the values of neighboring nodes and the value of the node itself as inputs. This gives rise to a Message Passing Neural Network (MPNN) architecture \cite{gilmer2017}. As we shall see in the following sections, our architecture is essentially an MPNN with communicating memory units whose graphical structure evolves in time. This kind of architecture, a GNN with communicating memory units, has been applied to the modeling of spatially interacting mobile agents in~\cite{casas2020}. GNNs in general have recently found use in multi-object tracking~\cite{weng2020,wang2020,li2020graph}; these models however, do not have the ability to pass messages to other parts in graph, and simply use the graph structure to infer similarity scores between consecutive sets of detections. We know of only one other approach which applies GNNs with message passing to the multi-object tracking problem~\cite{braso2020learning}. Unlike \cite{braso2020learning} where the model is run on batches of detections in an offline manner (with score averaging), the proposed approach involves dynamic graphs which are processed in a rolling-window scheme - making it suitable for online, real-time applications. Moreover, our message passing operations are specifically crafted with rapidly evolving graph structures in mind. 

The proposed approach is also partly motivated by the probabilistic graph representation of classical multi-target tracking methods, for example hypothesis based and track-tree based MHT (Multi-Hypothesis Tracking)~\cite{cox1996}\cite{reid1979}. In the these methods, the association of observations to targets/tracks is resolved by generating branches emanating from existing global hypotheses to account for new observations, then propagating the posterior probability of resulting global  hypotheses, recursively, along those branches. As in the earlier hypothesis based classical approaches, we do not generate explicit track-trees representing a set of underlying targets and we limit the size of the active section of the graph determining the associations to prevent exponential growth of compute. 
We also make no explicit assumptions about the underlying Bayesian distributions. The network learns from the ground truth tracking data, how best to generate the tracks during inference. Also, we do not enumerate the set of consistent tracks. The graph represents the superposition of all such tracks which can be decoded to produce the optimal one.

\section{Proposed Tracking Framework}\label{sec:framework}

This approach works based on the following idea: as data association in multi-object tracking (MOT) is conventionally modelled as a bipartite graph over two consecutive timesteps, could we simply learn and infer directly on such graph structures using GNNs? This graph structure could also be expanded to cover multiple consecutive timesteps so as to incorporate non-local information and infer distant associations.

Our approach formulates the MOT problem as inference on an undirected graph where each detection is represented as a node in the graph, and potential associations between different detections are represented by edges connecting them. The graph is dynamically updated at every new timestep, where new detections (and associations) from the latest timestep are added, and old, inactive detections (and associations) are removed. An overview of this setup is illustrated in Figure~\ref{fig:overview}. The proposed model (described in Section~\ref{sec:model}) works by operating on this dynamic graph in a rolling window basis (from left to right as presented in Figure~\ref{fig:overview}). The size of the rolling window is a design choice to be made based on the desired performance, available memory and compute requirements. This concept of a rolling window is similar to the one used in \cite{choi2015near}. 

Although not explicitly drawn in Figure~\ref{fig:overview} in order to reduce clutter, each edge joining two detection nodes is also treated as a node of the graph in actuality. This helps us endow a vector representation to each potential association between two detections. Thus, this dynamic graph structure is bipartite, with \textit{detection nodes} and \textit{association nodes} forming the two disjoint and independent sets. Detection nodes represent object detections in a sequence, and are initialized with their corresponding feature descriptors. Association nodes represent potential pairwise associations between two detections from different timesteps, and are initialized with zero vectors. These initializations are then transformed and updated by the model to create hidden representations of every node in the graph with each additional timestep.



\subsection{Training \& Inference}

\begin{algorithm*}[t]
\small
\caption{Pseudo-code for one training iteration}\label{alg:training}
\hspace*{\algorithmicindent} \textbf{Input:} \textit{feats, labels} 
\begin{algorithmic}[h]
\State $net \gets TrackMPNN()$ \Comment{initialize TrackMPNN model}
\State $G \gets initialize\_graph(feats)$ \Comment{initialize graph with detection features and pairwise associations from first two timesteps}
\State $total\_loss \gets 0$ \Comment{initialize loss to 0}
\State $(probabilities, loss) \gets net.forward(G, labels)$ \Comment{forward pass}
\State $total\_loss \gets total\_loss + loss$ \Comment{add to total loss}
\For{$t \gets 2$ \textbf{to} $t_{end}$}
\State $G \gets update\_graph(G, feats, t)$ \Comment{add new nodes and edges to graph; remove earliest nodes and edges}
\State $(probabilities, loss) \gets net.forward(G, labels)$ \Comment{forward pass}
\State $total\_loss \gets total\_loss + loss$ \Comment{add to total loss}
\If{condition}
\State $G \gets prune\_graph(G, probabilities)$ \Comment{remove low probability nodes to limit memory footprint}
\EndIf
\EndFor
\State $net \gets net.backward(total\_loss)$ \Comment{backward pass to train model}
\State \textbf{return} $net$
\end{algorithmic}
\end{algorithm*}

\begin{algorithm*}[t]
\small
\caption{Pseudo-code for inference on a MOT sequence}\label{alg:inference}
\hspace*{\algorithmicindent} \textbf{Input:} \textit{feats} 
\begin{algorithmic}[h]
\State $tracks \gets \{\}$ \Comment{initialize tracks to empty}
\State $net \gets TrackMPNN(trained\_weights)$ \Comment{initialize TrackMPNN model with trained weights}
\State $G \gets initialize\_graph(feats)$ \Comment{initialize graph with detection features and pairwise associations from first two timesteps}
\State $probabilities \gets net(G)$ \Comment{forward pass to get probabilities}
\State $tracks \gets tracks \bigcup decode\_graph(G, probabilities)$ \Comment{decode model probabilities to produce tracks for desired window}
\For{$t \gets 2$ \textbf{to} $t_{end}$}
\State $G \gets update\_graph(G, feats, t)$ \Comment{add new nodes and edges to graph; remove earliest nodes and edges}
\State $probabilities \gets net(G)$ \Comment{forward pass to get probabilities}
\State $tracks \gets tracks \bigcup decode\_graph(G, probabilities)$ \Comment{decode model probabilities to produce tracks for desired window}
\If{condition}
\State $G \gets prune\_graph(G, probabilities)$ \Comment{remove low probability nodes to limit memory footprint}
\EndIf
\EndFor
\State \textbf{return} $tracks$
\end{algorithmic}
\end{algorithm*}

Before we describe the model architecture and optimization in detail (Section~\ref{sec:model}), we provide a macro view of the training and inference procedure that we adopt. We make use of the following high level functions pertaining to the graph structure and the model during both training and inference:

\begin{itemize}
    \item \textit{initialize\_graph()}: This creates an initial bipartite graph with detection nodes from two consecutive timesteps, with an association node between every detection pair.
    \item \textit{update\_graph()}: This function is called at the start of every new timestep, to add new (detection and association) nodes and edges to end of the currently active graph. Note that association nodes are only added between the newly introduced detection nodes and unpaired detection nodes from previous timesteps. This function also removes the oldest set of nodes and edges from the currently active part of the graph. It essentially moves the rolling window one step forward.
    \item \textit{prune\_graph()}: This function removes low probability edges and nodes from the currently active part of the graph using a user specified confidence threshold. This function can be called whenever memory/compute requirements exceed what is permissible, or to prevent an explosion of nodes.
    \item \textit{decode\_graph()}: This function is called to decode the model-produced output probabilities for every node in the graph into corresponding object tracks. This can either be done in a greedy manner (by following the highest probability path from left to right) or by using the Hungarian algorithm (on consecutive timesteps from left to right).
    \item \textit{TrackMPNN()}: This initializes an instance of the proposed model (described in Section~\ref{sec:model}).
    \item \textit{TrackMPNN.forward()}: This carries out one forward pass of the data through the model.
    \item \textit{TrackMPNN.backward()}: This carries out one backward pass through the model to produce gradients with respect to the losses.
\end{itemize}

The training and inference pseudo-code that make use of these functions are presented in Algorithms~\ref{alg:training} and \ref{alg:inference}. As described earlier, the inference procedure operates in a rolling window manner, from past (left) to future (right). Thus, we operate on the entire tracking sequence in a continuous manner during inference. However, this is not possible during training, where gradients need to be stored for all detection and association nodes encountered in the dynamic graph. To make the training process computationally tractable, we split each tracking sequence into smaller contiguous chunks, and process each such mini-sequence as an individual sample. As depicted in Algorithm~\ref{alg:training}, we also accumulate losses over the sequence length and backpropagate only after the entire sequence has been processed.

\section{Model Architecture}\label{sec:model}

In this study, we use a class of Graph Neural Networks called Message Passing Neural Networks (MPNNs)~\cite{gilmer2017}. Like most MPNNs, our proposed TrackMPNN model consists of:
\begin{itemize}
    \item a message function:
    \begin{equation}
        \vv{m}_v^{k} = \sum_{w \in \mathcal{N}(v)} M(\vv{h}_v^{k-1}, \vv{h}_w^{k-1}),
    \end{equation}
    
    \item a node/vertex update function:
    \begin{equation}
        \vv{h}_v^{k} = U(\vv{h}_v^{k-1}, \vv{m}_v^{k}),
    \end{equation}
    
    \item a readout/output function:
    \begin{equation}
        \hat{y}_v^{k} = R(\vv{h}_v^{k}),
    \end{equation}
\end{itemize}
where vertices $v, w$ are nodes in graph $G$, and in our context be either detection nodes or association nodes.
Note that iteration $k$ is different from the current timestep $t$ of the tracking sequence. Whereas $t$ signifies the lifetime of the entire graph, $k$ denotes the lifetime of each individual node in this dynamic graph.
For our purposes, we use separate functions/weights for detection nodes and association nodes, which we detail in the next two subsections. 

\subsection{Detection node operations}\label{sec:det_node}

Consider the illustration of a detection node $d$ and its neighboring association nodes $a_i \in \mathcal{N}(d)$ presented in Figure~\ref{fig:det_node}. The operations associated with nodes of this type are presented below.

\begin{figure}[h]
    \center{\includegraphics[width=0.8\linewidth]
    {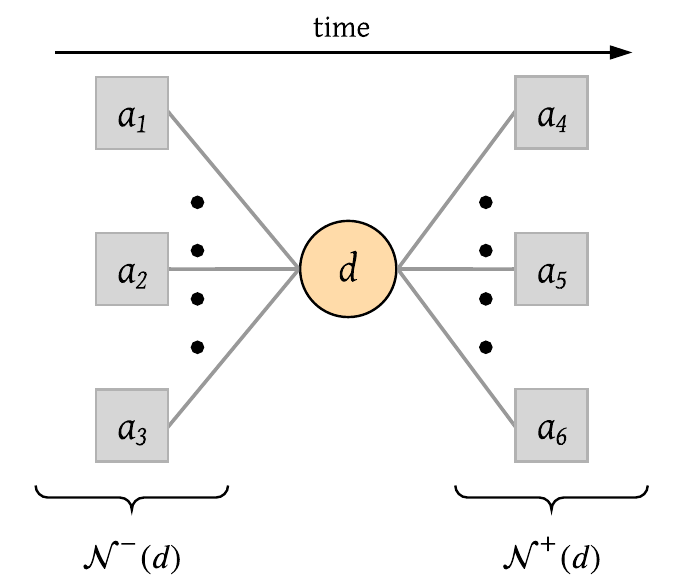}}
    \caption{\label{fig:det_node} Illustration of a detection node $d$ and its neighboring association nodes $\{a_i | {a_i \in \mathcal{N}(d)}\}$. The neighborhood can be split into two disjoint sets $\mathcal{N}^-(d)$ and $\mathcal{N}^+(d)$, denoting all nodes from the past and future respectively.}
\end{figure}

\subsubsection{Initialization}\label{sec:det_node_init}
The hidden state of a detection node is initialized with a pair of linear transformations (with non-linearity and normalization) of its input feature representation $\vv{x}_d$:
\begin{equation}\label{eq:det_init_0}
    \vv{h}_d^0 = \mathbf{W}_{det}^{i'}\ \text{BatchNorm}(\vv{x'}_d) + \vv{b}_{det}^{i'},
\end{equation}
where
\begin{equation}\label{eq:det_init_1}
    \vv{x'}_d = \text{ReLU}(\mathbf{W}_{det}^i \vv{x}_d + \vv{b}_{det}^i).
\end{equation}

The input representation $\vv{x}_d$ of a detection node $d$ can be composed of different features and attributes describing the detected object instance (a detection produced by the object detector). In this study, we use detections produced by a recurrent rolling convolutions (RRC) detector~\cite{ren2017accurate} - though any other detector could be used instead. Contrary to many contemporary approaches, we only use the 2D box locations and object categories produced by the detector to describe each object instance, and rely on the model to learn the patterns of sequential locations on the image plane. 

The input representation of each detection $\vv{x}_d$ is then defined as follows:
\begin{equation}~\label{eq:feats}
    \vv{x}_d = [\vv{x}_{d;2d} || \vv{x}_{d;cat}],
\end{equation}
where $||$ represents concatenation along the singleton dimension.
\begin{equation}
    \vv{x}_{d;2d} = (x_1, y_1, x_2 - x_1, y_2 - y_1, score)^\intercal
\end{equation} 
denotes 2D features comprised of the 2D bounding box location (top-left location, width, height) and score, and,
\begin{equation}
    \vv{x}_{d;cat} = (0, \cdots, 1, \cdots, 0)^\intercal
\end{equation}
is the one-hot encoding of the object category.

\subsubsection{Message and node/vertex update functions}~\label{sec:node_update}
The hidden state of the detection node is updated based on its previous hidden state and the previous hidden states of its neighboring association nodes, weighted by an attention mechanism that takes into account the detection nodes connected by the association:
\begin{equation}
\begin{split}\label{eq:node-update}
    \vv{h}_d^k &= \text{GRU}\big(\vv{h}_d^{k-1}, \vv{m}_{d}^{k}\big) \\
    &= \text{GRU}\bigg(\vv{h}_d^{k-1}, \sum_{a_i \in \mathcal{N}(d)} \alpha_{da_i}^{k-1} \vv{h}_{a_i}^{k-1}\bigg),
\end{split}
\end{equation}
where $\vv{m}_{d}^{k}$ is the message passed to detection node $d$ at iteration $k$, and $\alpha_{da_i}^{k-1}$ are the attention weights~\cite{velivckovic2017graph} assigned to each association node $a_i^{k-1}$. 

Fully expanded out, the coefficients computed by the attention mechanism may then be expressed as:
\begin{equation}\label{eq:attention}
\resizebox{.9\hsize}{!}{
    $\alpha_{da_i}^{k-1} = \frac{exp(LReLU(\mathbf{a}^\intercal|\mathbf{W^h_{det}}\vv{h}_d^{k-1} - \mathbf{W^h_{det}}\vv{h}_{d_i}^{k-1}|))}{\sum_{d_j \in \mathcal{N}^2(d)} exp(LReLU(\mathbf{a}^\intercal|\mathbf{W^h_{det}}\vv{h}_d^{k-1} - \mathbf{W^h_{det}}\vv{h}_{d_j}^{k-1}|))}$
},
\end{equation}
where $d_i \in \mathcal{N}(a_i)$ and $d_i \neq d$, $\mathcal{N}^2(\cdot)$ represents the second-order neighborhood of a node, and $\text{LReLU}$ denotes the LeakyReLU non-linearity (with negative input slope $0.2$). 

\subsubsection{Readout function}
The detection node output $o_d^k$ represents a scalar confidence value obtained by simple linear transformation of its hidden state:
\begin{equation}\label{eq:det_readout}
    o_d^k = \mathbf{W}_{det}^{o} \vv{h}_d^k + \vv{b}_{det}^o.
\end{equation}

\subsection{Association node operations}\label{sec:ass_node}

Consider the illustration of an association node $a$ and its neighboring detection nodes depicted in Figure~\ref{fig:ass_node}. The operations associated with nodes of this type are presented below.

\begin{figure}[h]
    \center{\includegraphics[width=0.35\textwidth]
    {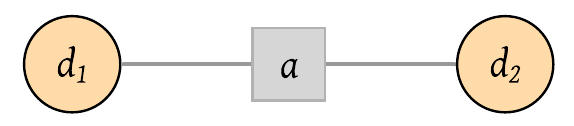}}
    \caption{\label{fig:ass_node} Illustration of an association node $a$ and its two neighboring detection nodes $\{d_1, d_2\}$.}
\end{figure}

\noindent
\subsubsection{Initialization} 
The hidden state of an association node is initialized with a vector of of $0$s.
\begin{equation}
    \vv{h}_a^0 = \vv{0}
\end{equation}

\subsubsection{Message and node/vertex update functions}
The hidden state of the association node is updated based on its previous hidden state and the previous hidden states of its two neighboring detection nodes. Given the fixed degree of an association node, we experiment with two different message update functions. The first is based on the difference between the hidden states of its neighbors: 
\begin{equation}\label{eq:diff}
\begin{split}
    \vv{h}_a^k &= \text{GRU}\big(\vv{h}_a^{k-1}, \vv{m}_{a}^{k}\big) \\
    &= \text{GRU}\bigg(\vv{h}_a^{k-1}, \mathbf{W}_{ass}^h [\vv{h}_{d_2}^{k-1} - \vv{h}_{d_1}^{k-1}] + \vv{b}_{ass}^h\bigg),
\end{split}
\end{equation}
and the second is based on their concatenation:
\begin{equation}\label{eq:concat}
\begin{split}
    \vv{h}_a^k &= \text{GRU}\big(\vv{h}_a^{k-1}, \vv{m}_{a}^{k}\big) \\
    &= \text{GRU}\bigg(\vv{h}_a^{k-1}, \mathbf{W}_{ass}^h [\vv{h}_{d_1}^{k-1} || \vv{h}_{d_2}^{k-1}] + \vv{b}_{ass}^h\bigg).
\end{split}
\end{equation}
In both cases, $\vv{m}_{a}^{k}$ is the message passed to association node $a$ at iteration $k$.

\subsubsection{Readout function}
The association node output $o_a^k$ represents a scalar confidence value obtained by simple linear transformation of its hidden state:
\begin{equation}\label{eq:ass_readout}
    o_a^k = \mathbf{W}_{ass}^{o} \vv{h}_a^k + \vv{b}_{ass}^o.
\end{equation}

\subsection{Training Losses}

Let $G = (V, E)$ denote the dynamic graph at any given instant. We can further split the set of vertices into two disjoint sets, i.e. $V = DN \cup AN$ and $DN \cap AN = \emptyset$, where $DN$ and $AN$ represent the set of detection nodes and association nodes respectively.

We first apply a binary cross-entropy loss at each detection node:

\begin{multline}\label{eq:TP}
    \mathcal{L}_{bce; DN} = -\frac{1}{|DN|} \sum_{d \in DN} \bigg(y_d \log(\text{Sigmoid}(o_d^{k_d}))\\ 
    + (1 - y_d) \log(1 - \text{Sigmoid}(o_d^{k_d}))\bigg),
\end{multline}
where $k_d$ denotes the latest iteration of detection node $d$, and
\begin{equation}
    y_d = 
    \begin{cases}
    1, & \text{if true positive}\\
    0, & \text{otherwise.}
    \end{cases}
\end{equation}
Similarly, we apply a binary cross-entropy loss at each association node:
\begin{multline}
    \mathcal{L}_{bce; AN} = -\frac{1}{|AN|} \sum_{a \in AN} \bigg(y_a \log(\text{Sigmoid}(o_a^{k_a}))\\ 
    + (1 - y_a) \log(1 - \text{Sigmoid}(o_a^{k_a}))\bigg),
\end{multline}
where $k_a$ denotes the latest iteration of association node $a$, and
\begin{equation}
    y_a = 
    \begin{cases}
    1, & \text{if } \mathcal{N}(a) \text{ belongs to the same track}\\
    0, & \text{otherwise.}
    \end{cases}
\end{equation}
We also apply a cross-entropy loss for every set of competing association nodes:
\begin{multline}
    \mathcal{L}_{ce; AN} = -\frac{1}{|DN|} \sum_{d \in DN} \bigg(\\ 
    \frac{1}{|\mathcal{N}^{-}(d)|} \sum_{a \in \mathcal{N}^{-}(d)} y_a \log(\text{Softmax}(o_a^{k_a})) \\
    + \frac{1}{|\mathcal{N}^{+}(d)|} \sum_{a \in \mathcal{N}^{+}(d)} y_a \log(\text{Softmax}(o_a^{k_a})) \bigg),
\end{multline}
where
\begin{equation}
    y_a = 
    \begin{cases}
    1, & \text{if } \mathcal{N}(a) \text{ belongs to the same track}\\
    0, & \text{otherwise,}
    \end{cases}
\end{equation}
$\mathcal{N}^{-}(\cdot)$ is the set of all neighbors from past timesteps, and $\mathcal{N}^{+}(\cdot)$ is the set of all neighbors from future timesteps.

The total loss used to train the entire model is the sum of the individual losses:
\begin{equation}\label{eq:tot_loss}
    \mathcal{L} = \mathcal{L}_{bce;DN} + \mathcal{L}_{bce;AN} + \mathcal{L}_{ce;AN}
\end{equation}

\section{Experiments and Analyses}

\subsection{Implementation Details}
\noindent\textbf{Dynamic Undirected Graph}: As detailed in Section~\ref{sec:framework}, our approach operates using a temporal rolling window, where detection and association nodes falling inside the window make up the dynamic graph at any given instant. The two hyperparameters that govern this process are the current window size (CWS) and the retained window size (RWS). CWS defines the size of the rolling window in discrete timesteps/frames. RWS determines the number of discrete timesteps for which an unassociated detection node is kept in the dynamic graph after it is no longer in the rolling window. This is to ensure that objects that are temporarily occluded or missing have a chance to be re-identified again. The rolling window is updated after each timestep by adding new nodes from the next timestep and removing nodes from the earliest timestep. Before detection nodes are removed however, they are assigned to tracks as part of the decoding process i.e., for every detection node that is removed, it is assigned to the same track as an earlier detection node corresponding to the maximum association probability as produced by the model (Eq.~\ref{eq:ass_readout}). If a true positive detection node does not have any high probability associations, a new track is initialized.\\
\noindent\textbf{Training \& Optimization}: As depicted in Eq.~\ref{eq:tot_loss}, the total loss is simply the sum of the individual losses - without any weighting. We also forego mini-batch training and opt for \textit{mini-sequence} training in its place. Instead of using a mini-batch of examples for each training sample, we use one mini-sequence for each training sample. A mini-sequence is simply a tracking sequence of CWS contiguous timesteps, randomly sampled from full-length tracking sequences. Similar to mini-batch gradient descent, the losses are accumulated, and then backpropagated after the entire mini-sequence is processed. In our experience, this tends to stabilize training, produce better gradients, and require less memory. 
The entire network is trained using an Adam optimizer with learning rate $0.0001$, $\beta_1=0.9$ and $\beta_2=0.999$ for $50$ epochs. Additionally, we augment the training split with multiple random transformations. This includes time reversal (reverse timestep/frame ordering), dropout of detections (ignore a fraction of true positive detections), and horizontal flips. Finally, we initialize the bias values in the detection and association node readout functions (Eq.~\ref{eq:det_readout}, \ref{eq:ass_readout}) to $+4.595$ and $-4.595$ respectively. This is to keep the losses manageable during the initial phase of training. \\
\noindent\textbf{Inference}: During inference, care is taken to ensure that hyperparameters associated with the dynamic graph remain unchanged from training. Instead of feeding the model mini-sequences, we instead supply the full-length sequence as a single sample. The model operates on this sequence in a rolling window manner, generating continuous tracks along the way.

\begin{figure}[t]
\captionsetup[subfigure]{justification=centering}
  	\centering
  	\begin{subfigure}[t]{0.15\textwidth}
		\centering
		\includegraphics[width=\linewidth]{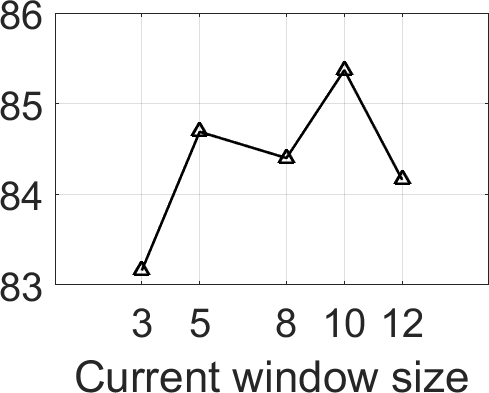}
		\caption{MOTA versus Current Window Size (RWS=0)}
		\label{fig:cws}
	\end{subfigure}%
~  	
	\begin{subfigure}[t]{0.15\textwidth}
		\centering
		\includegraphics[width=\linewidth]{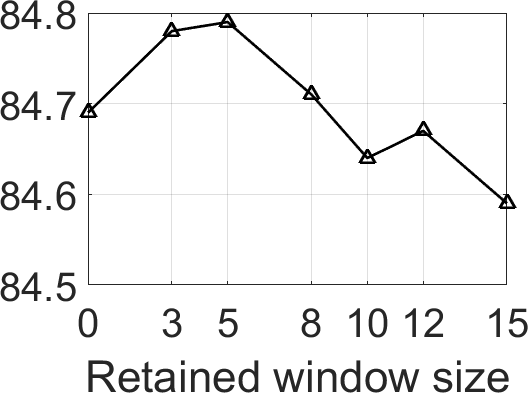}
		\caption{MOTA versus Retained Window Size (CWS=5)}
		\label{fig:rws}
	\end{subfigure}
~
	\begin{subfigure}[t]{0.15\textwidth}
		\centering
		\includegraphics[width=\linewidth]{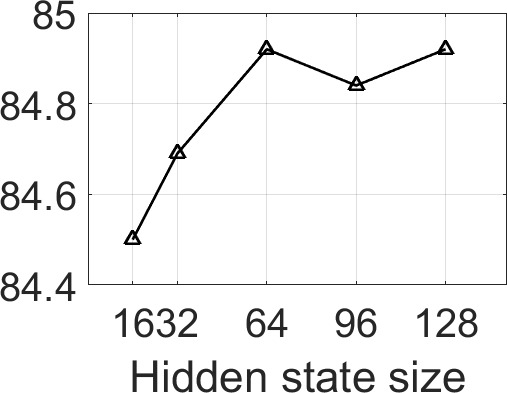}
		\caption{MOTA versus Hidden State Size (CWS=5, RWS=0)}
		\label{fig:hss}
	\end{subfigure}%
	\caption{Experiments illustrating the effect of different model settings on the tracking performance using the KITTI validation split (for Cars only).}
	\vspace{-3mm}
	\label{fig:val-exp}
\end{figure}

\subsection{Ablation Experiments}
To perform ablation experiments, we make use of the KITTI MOT dataset~\cite{Geiger2012CVPR}. The KITTI MOT dataset is comprised of 21 training sequences and 29 testing sequences, with two categories of interest: Cars and Pedestrians. For the purpose of ablation, we use the first 11 training sequences to train the models, and leave the rest for validation. We also restrict our experiments to one object category - Cars. For evaluation, we make use of the CLEAR MOT metrics~\cite{bernardin2008evaluating, li2009learning}. In cases where only one metric is desired, we use the Multi-Object Tracking Accuracy (MOTA). The following paragraphs detail our ablation of different settings pertaining to the dynamic graph and the model.\\
\noindent\textbf{Dynamic graph settings}: To assess the benefits of a potentially larger rolling window size, we trained multiple models with different settings of CWS. In all cases, RWS was set to $0$. Figure~\ref{fig:cws} depicts the plot of MOTA for different settings of CWS. Although all variants achieve very similar results, CWS=10 produces the best MOTA. CWS=5 achieves very similar results at a lower computational and memory cost. Next, we fix CWS=5 and evaluate models with different values of RWS. The results from this experiment are plotted in Figure~\ref{fig:rws}. Once again, we notice that the MOTA values are quite similar, with RWS=5 achieving the best results. Both these experiments also highlight the robustness of the model across various graph settings.\\
\noindent\textbf{Model settings}: To compare different model settings and their impacts on performance, we first define a baseline model (B) against which other variants can be compared. The baseline model is trained using graph settings CWS=5, RWS=0, and uses difference based updates for the association nodes (Eq.~\ref{eq:diff}). Results from these comparisons are presented in Table~\ref{tab:val-exp}. First, we measure the efficacy of the greedy matching procedure by comparing it with a model using the Hungarian algorithm for optimal bipartite matching (B w/ H). In this variant, the Hungarian algorithm is used to decode the graph into tracks, by optimally matching detections to existing tracks based on association probabilities. We notice that the results are better across the board. Thus, the Hungarian algorithm helps produce slightly longer and continuous tracks - at the expense of additional computational cost. Second, we train a model (B w/o TP) without true-positive classification at the detection nodes (see Eq.~\ref{eq:TP}). In this case, every detection is assumed to be a true positive, and the model is used to only produce association probabilities. From the results presented in Table~\ref{fig:val-exp}, we see that this setting also improves performance on all metrics. This makes sense because 2D box locations alone are usually not enough to distinguish true positives from false positives. Third, to compare the two different update functions for association nodes, we also train a variant of the baseline with concatenation based updates (Eq.~\ref{eq:concat}). When compared to the difference based updates in B, concatenation produces worse results with a much lower MOTA and a higher IDS. This indicates that difference based updates can better model the similarity/dissimilarity between two detection nodes. Next, to quantify the effects of data augmentation during training, we train a model with random transformations (B w/ RT). This too improves most metrics across the board - indicating the clear benefits of our augmentation scheme  - especially when trained on datasets of limited size. 
Finally, to gauge the effects of hidden state size on the tracking performance, we train multiple models containing GRUs with different hidden state sizes. The resulting MOTAs for different variants are plotted in Figure~\ref{fig:hss}. The plot indicates that increasing the dimensionality of the hidden state tends to improve performance, but this trend saturates beyond $64$ dimensions. 

\begin{table}[t]
\centering
\resizebox{0.95\linewidth}{!}{%
\begin{threeparttable}\centering
\caption{Results from experiments on the KITTI validation set (for Cars only). Arrows indicate if a higher ($\uparrow$) or lower value of the metric ($\downarrow$) is desired.}
\label{tab:val-exp}
\begin{tabular}{@{}ccccccc@{}}
\hline
\begin{tabular}[c]{@{}c@{}}Model type\end{tabular} & 
\begin{tabular}[c]{@{}c@{}}MOTA\\ ($\uparrow$)\end{tabular} & 
\begin{tabular}[c]{@{}c@{}}MOTP\\ ($\uparrow$)\end{tabular} & 
\begin{tabular}[c]{@{}c@{}}MT\\ ($\uparrow$)\end{tabular} & 
\begin{tabular}[c]{@{}c@{}}ML\\ ($\downarrow$)\end{tabular} & 
\begin{tabular}[c]{@{}c@{}}IDS\\ ($\downarrow$)\end{tabular} &
\begin{tabular}[c]{@{}c@{}}FRAG\\ ($\downarrow$)\end{tabular}\\
\hline \hline
\rowcolor[HTML]{EFEFEF}
B\tnote{1}                 & 84.69 & 0.1009 & 71.49\% & 2.26\% & 320 & 156 \\
B w/ H\tnote{2}            & 85.99 & 0.1012 & 73.20\% & 2.26\% & 140 & 152 \\
\rowcolor[HTML]{EFEFEF}
B w/o TP\tnote{3}          & 85.18 & 0.1013 & 72.85\% & 2.26\% & 239 & 153 \\
B$||$\tnote{4}          & 79.30 & 0.1108 & 65.71\% & 5.61\% & 703 & 205 \\
\rowcolor[HTML]{EFEFEF}
B w/ RT\tnote{5}          & 84.86 & 0.1008 & 72.85\% & 2.26\% & 291 & 153 \\
\bottomrule
\end{tabular}
\begin{tablenotes}
    \item[1] B: baseline model with difference based updates for association nodes
    \item[2] H: Hungarian algorithm
    \item[3] TP: true positive classification
    \item[4] B$||$: baseline model with concatenation based updates for association nodes (Eq.~\ref{eq:concat})
    \item[5] RT: random transformations during training
 \end{tablenotes}
\end{threeparttable}
}
\end{table}

\begin{table*}[t]
\centering
\resizebox{0.8\linewidth}{!}{%
\begin{threeparttable}\centering
\caption{Results on the KITTI multi-object tracking benchmark for Cars. Arrows indicate if a higher ($\uparrow$) or lower value of the metric ($\downarrow$) is desired. Our results are in bold.}
\label{tab:kitti-cars}
\begin{tabular}{@{}ccccccccc@{}}
\hline
\begin{tabular}[c]{@{}c@{}}Method\end{tabular} & 
\begin{tabular}[c]{@{}c@{}}Sensors \end{tabular} & 
\begin{tabular}[c]{@{}c@{}}Online \end{tabular} & 
\begin{tabular}[c]{@{}c@{}}MOTA\\ ($\uparrow$)\end{tabular} & 
\begin{tabular}[c]{@{}c@{}}MOTP\\ ($\uparrow$)\end{tabular} & 
\begin{tabular}[c]{@{}c@{}}MT\\ ($\uparrow$)\end{tabular} & 
\begin{tabular}[c]{@{}c@{}}ML\\ ($\downarrow$)\end{tabular} & 
\begin{tabular}[c]{@{}c@{}}IDS\\ ($\downarrow$)\end{tabular} &
\begin{tabular}[c]{@{}c@{}}FRAG\\ ($\downarrow$)\end{tabular}\\
\hline \hline
\rowcolor[HTML]{EFEFEF}
CenterTrack \cite{zhou2020tracking} & RGB camera & \ding{51}\tnote{1} & 0.8883 & 0.8497 & 82.15 \% & 2.46 \% & 254 & 227 \\
\textbf{TrackMPNN + CenterTrack} & \textbf{RGB camera} & \ding{51} & \textbf{0.8733} & \textbf{0.8449} & \textbf{84.46 \%} & \textbf{2.15 \%} & \textbf{481} & \textbf{237} \\
\hline
JRMOT \cite{Shenoi2020JRMOTAR} & RGB camera, LiDAR & \ding{51} & 0.8510 & 0.8528 & 70.92 \% & 4.62 \% & 271 & 273 \\
\rowcolor[HTML]{EFEFEF}
\textbf{TrackMPNN + RRC} & \textbf{RGB camera} & \ding{51} & \textbf{0.8455} & \textbf{0.8507} & \textbf{70.92 \%} & \textbf{4.00 \%} & \textbf{466} & \textbf{482} \\
mono3DT \cite{Hu3DT19} & RGB camera, GPS & \ding{51} & 0.8428 & 0.8545 & 73.08 \% & 2.92 \% & 379 & 573 \\
\rowcolor[HTML]{EFEFEF}
MOTSFusion \cite{luiten2019MOTSFusion} & RGB camera, stereo & \ding{51} & 0.8424 & 0.8503 & 72.77 \% & 2.92 \% & 415 & 569 \\
\rowcolor[HTML]{EFEFEF}
SMAT \cite{10.1007/978-3-030-50516-5_5} & RGB camera & \ding{51} & 0.8364 & 0.8589 & 62.77 \% & 6.00 \% & 198 & 294 \\
mmMOT \cite{mmMOT_2019_ICCV} & RGB camera & \ding{53}\tnote{2} & 0.8323 & 0.8503 & 72.92 \% & 2.92 \% & 733 & 570 \\
\rowcolor[HTML]{EFEFEF}
MOTBeyondPixels \cite{MOTBeyondPixels} & RGB camera & \ding{51} & 0.8268 & 0.8550 & 72.61 \% & 2.92 \% & 934 & 581 \\
\bottomrule
\end{tabular}
\begin{tablenotes}
    \item[1] online/near-online
    \item[2] offline/batch
 \end{tablenotes}
\end{threeparttable}
}
\end{table*}

\begin{table*}[t]
\centering
\resizebox{0.8\linewidth}{!}{%
\begin{threeparttable}\centering
\caption{Results on the KITTI multi-object tracking benchmark for Pedestrians. Arrows indicate if a higher ($\uparrow$) or lower value of the metric ($\downarrow$) is desired. Our results are in bold.}
\label{tab:kitti-peds}
\begin{tabular}{@{}ccccccccc@{}}
\hline
\begin{tabular}[c]{@{}c@{}}Method\end{tabular} & 
\begin{tabular}[c]{@{}c@{}}Sensors \end{tabular} & 
\begin{tabular}[c]{@{}c@{}}Online \end{tabular} & 
\begin{tabular}[c]{@{}c@{}}MOTA\\ ($\uparrow$)\end{tabular} & 
\begin{tabular}[c]{@{}c@{}}MOTP\\ ($\uparrow$)\end{tabular} & 
\begin{tabular}[c]{@{}c@{}}MT\\ ($\uparrow$)\end{tabular} & 
\begin{tabular}[c]{@{}c@{}}ML\\ ($\downarrow$)\end{tabular} & 
\begin{tabular}[c]{@{}c@{}}IDS\\ ($\downarrow$)\end{tabular} &
\begin{tabular}[c]{@{}c@{}}FRAG\\ ($\downarrow$)\end{tabular}\\
\hline \hline
\rowcolor[HTML]{EFEFEF}
CenterTrack \cite{zhou2020tracking} & RGB camera & \ding{51} & 0.5384 & 0.7372 & 35.40 \% & 21.31 \% & 425 & 618 \\
\textbf{TrackMPNN + CenterTrack} & \textbf{RGB camera} & \ding{51} & \textbf{0.5210} & \textbf{0.7342} & \textbf{35.05 \%} & \textbf{18.90 \%} & \textbf{626} & \textbf{669} \\
\bottomrule
\end{tabular}
\begin{tablenotes}
    \item[1] online/near-online
    \item[2] offline/batch
 \end{tablenotes}
\end{threeparttable}
}
\end{table*}

\begin{figure*}[t]
    \center{\includegraphics[width=0.9\linewidth]{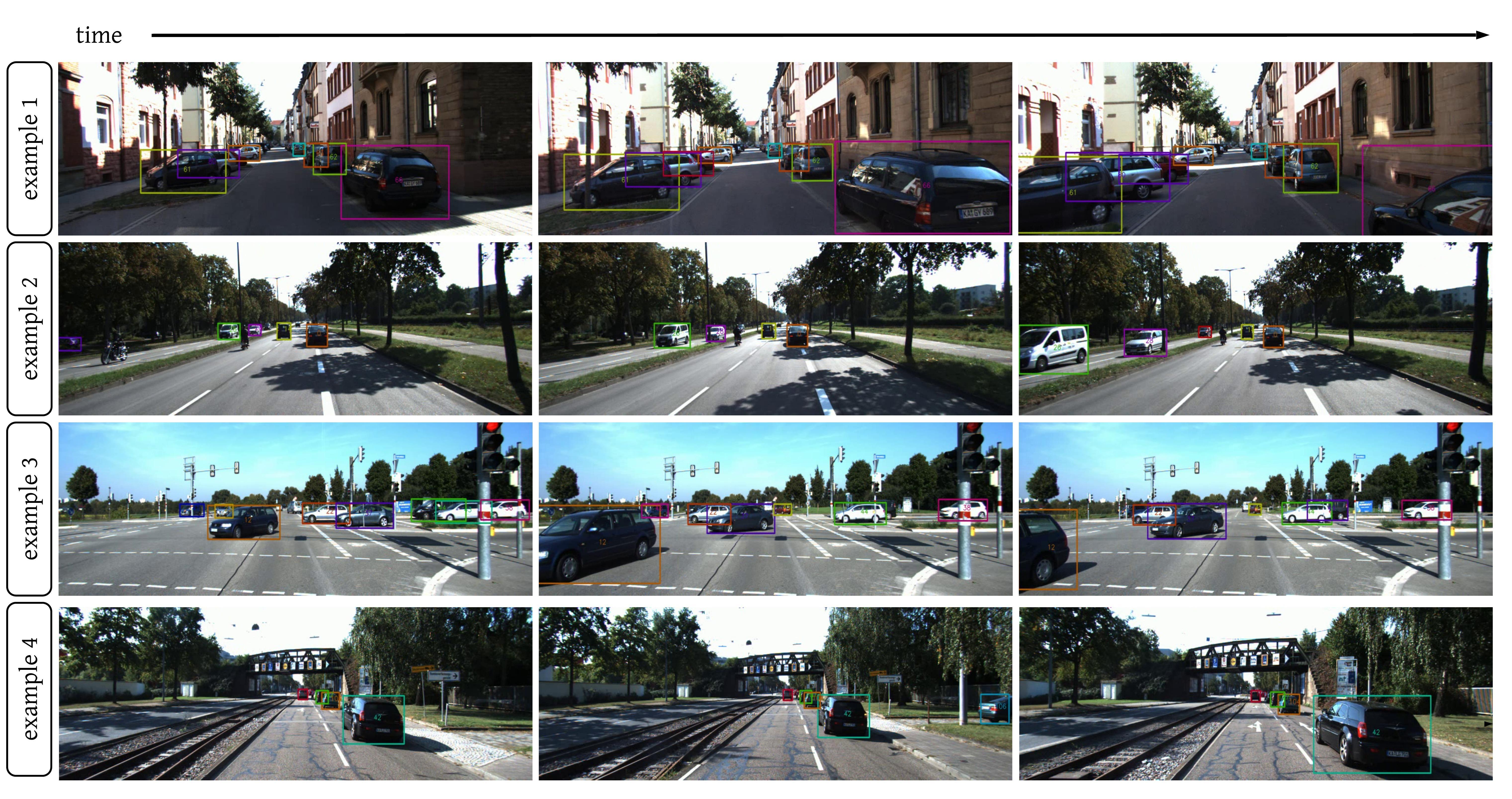}}
    \caption{Qualitative examples on the KITTI multi-object tracking (MOT) testing set. Each row depicts a sequence of frames, with overlaid color-coded detection boxes. Each color represents a unique track generated by our model.}
    \label{fig:kitti_qual}
\end{figure*}

\subsection{Comparison with State of the Art}\label{sec:state-of-art}
\noindent\textbf{Model and graph settings}: To compare our proposed approach with existing methods, we train a TrackMPNN model with CWS=5, and incorporate random transformations for data augmentation. We also discard true positive classification, and use the Hungarian algorithm during inference.
This model is trained with the same optimizer settings as the baseline for a total of $30$ epochs, using $18$ of the $21$ training sequences provided in KITTI MOT. The remaining $3$ sequences were used for validation and early stopping.\\
\noindent\textbf{KITTI MOT benchmark}: We compare two variants of our approach with other competing methods on the KITTI MOT benchmark in Table~\ref{tab:kitti-cars}. One variant makes use of RRC detections~\cite{ren2017accurate} (TrackMPNN + RRC), and the other uses detections produced by the CenterTrack tracker~\cite{zhou2020tracking} (TrackMPNN + CenterTrack). Our TrackMPNN + RRC model is trained to track Cars, whereas TrackMPNN + CenterTrack is trained to track all object categories on KITTI i.e., Cars, Pedestrians  and Cyclists. First, our TrackMPNN + RRC model outperforms all competing 2D methods that make use of the same set of detections~\cite{karunasekera2019multiple, Hu3DT19, luiten2019MOTSFusion, 10.1007/978-3-030-50516-5_5, mmMOT_2019_ICCV, MOTBeyondPixels} on the MOTA metric, and remains competitive on other metrics. For RRC detections, we only fall short of JRMOT~\cite{Shenoi2020JRMOTAR} which makes use of LiDAR point clouds for 3D range information.
Next, when using the same set of detections as CenterTrack, we beat every other approach and are comparable in terms of performance to the state-of-the-art. Once again, we do this without relying on any visual cues or pre-training on other datasets as in~\cite{zhou2020tracking}. The performance of our TrackMPNN + CenterTrack model is also comparable to~\cite{zhou2020tracking} for tracking Pedestrians (see Table~\ref{tab:kitti-peds}) when using the same set of detections. While we fall short of CenterTrack on a few metrics, we beat them in some others (MT and ML).\\
\noindent\textbf{Qualitative results from KITTI MOT}: In addition to the quantitative results presented above, we also show some qualitative results of our method from the KITTI MOT testing set in Figure~\ref{fig:kitti_qual}. The results depict multiple examples of tracking across frames in a sequence, overlaid with detection boxes. The boxes are color coded to indicate each unique track produced by our model.

\subsection{Analyzing Learned Attention Weights}~\label{sec:att}
As described in Section~\ref{sec:node_update} of the paper, we make use of graph attention for detection nodes $d$ to selectively receive messages from neighboring association nodes  $\{a_i | a_i \in \mathcal{N}(d)\}$. However, we do not enforce any restrictions on which associations in the neighborhood should receive more/less weights. This is learned by the model when optimizing for the losses during training.

To better understand the nature of these attention weights assigned by the model, we train a baseline model $B$ with multi-head attention comprised of three attention heads. We use the same settings and train-val split as the ablation experiments. The trained model is then run on the validation set, and the attention weights assigned to every association node are grouped based on whether the underlying association is true or false, i.e. if the neighboring detections connected by the association node belong to the same track or not. 

\begin{figure}[t]
    \center{\includegraphics[width=0.9\linewidth]{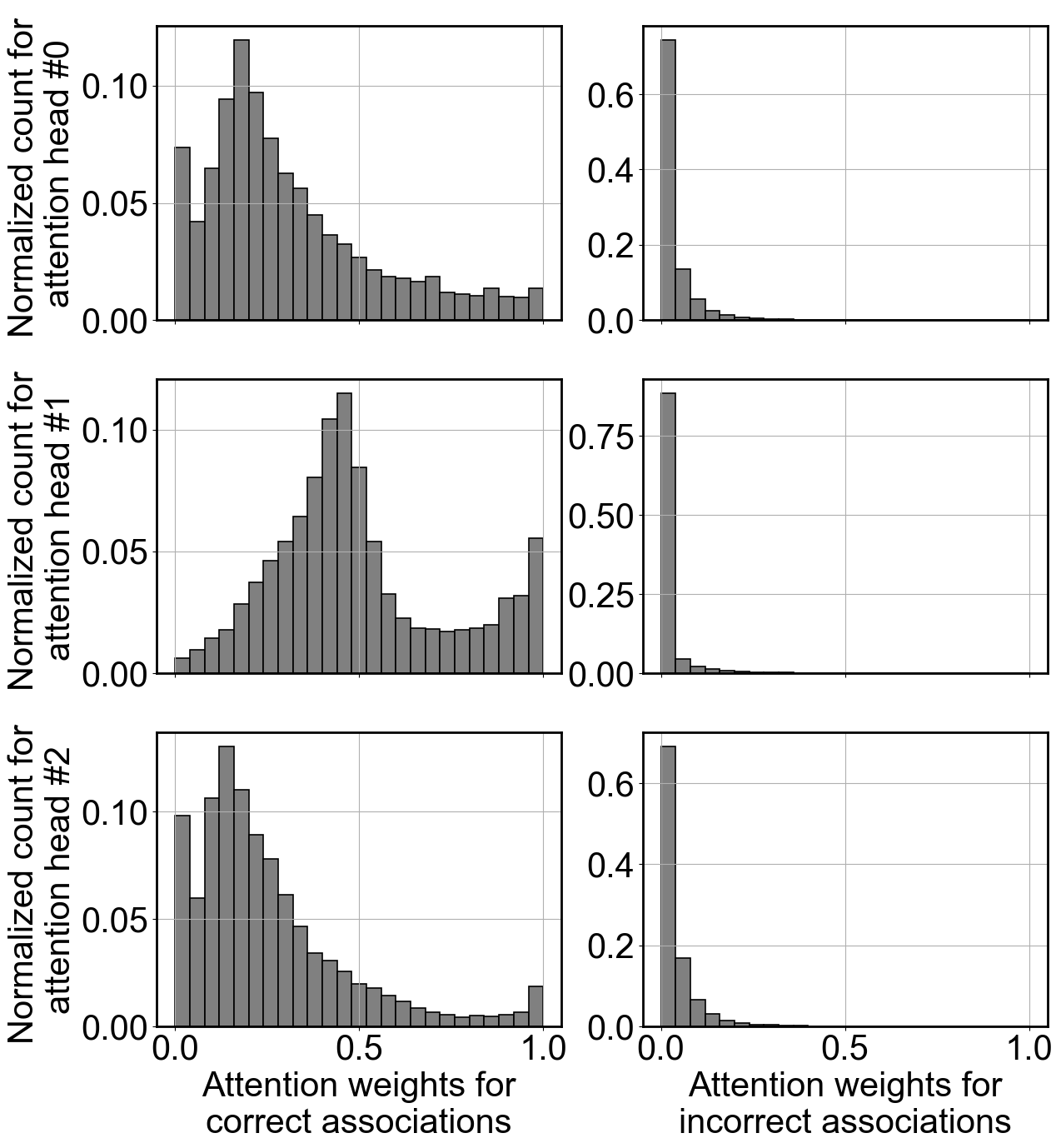}}
    \caption{Histogram plots of the attention weights assigned to true (left column) and false associations (right column) for each of three attention heads. Each row above corresponds to one of three attention heads.}
    \label{fig:att_dist}
\end{figure}

The distributions of these two separate sets of attention weights (for each of three attention heads) are presented in Figure~\ref{fig:att_dist}. First, we see that higher attention weights are generally assigned to true associations, while those for incorrect associations are skewed towards 0. This implies that messages are mostly exchanged between detection nodes belonging to the same track. Next, we notice different distributions for each attention head, indicating that each head tends to receive different messages from the neighborhood. This is especially true for the second attention head in Figure~\ref{fig:att_dist}. This experiment clearly indicates that the model learns to weight the messages appropriately without any explicit supervision.

\begin{figure*}[ht]
\captionsetup[subfigure]{justification=centering}
  	\centering
  	\begin{subfigure}[t]{0.24\textwidth}
		\centering
		\includegraphics[width=\linewidth]{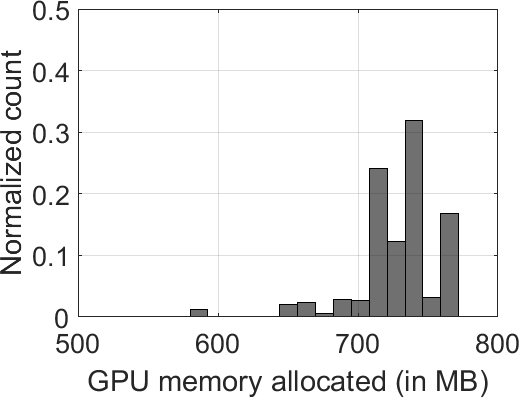}
		\caption{Memory allocated during training}
		\label{fig:train-mem-alloc}
	\end{subfigure}%
~  	
	\begin{subfigure}[t]{0.24\textwidth}
		\centering
		\includegraphics[width=\linewidth]{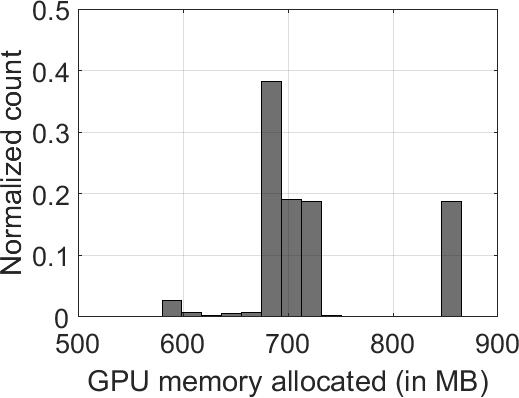}
		\caption{Memory allocated during inference}
		\label{fig:val-mem-alloc}
	\end{subfigure}
~
	\begin{subfigure}[t]{0.24\textwidth}
		\centering
		\includegraphics[width=\linewidth]{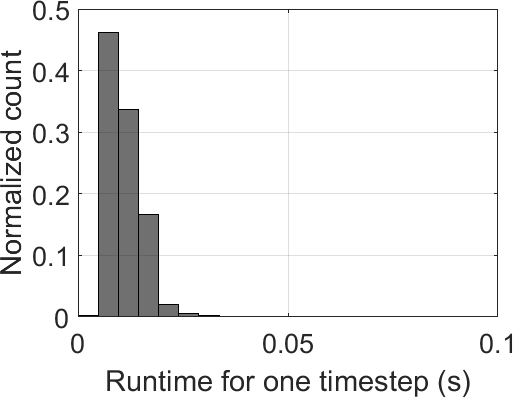}
		\caption{Inference runtime}
		\label{fig:infer-runtime}
	\end{subfigure}%
~  	
	\begin{subfigure}[t]{0.24\textwidth}
		\centering
		\includegraphics[width=\linewidth]{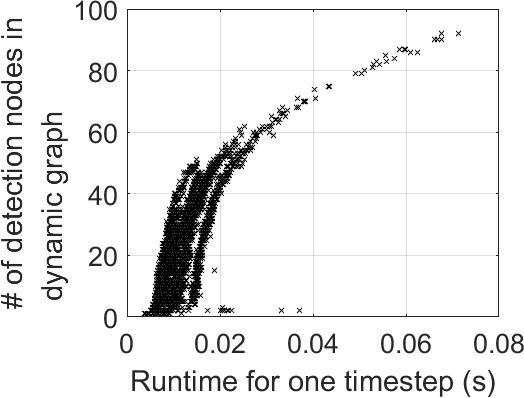}
		\caption{Number of detection nodes versus runtime}
		\label{fig:objects-vs-runtime}
	\end{subfigure}
	\caption{Plots examining the compute requirements and runtime of our proposed approach when using RRC detections on the KITTI MOT dataset.}
	\vspace{-3mm}
	\label{fig:mem-alloc}
\end{figure*}
\subsection{Compute Requirements and Runtime}
Since this approach relies on dynamic graphs and operations on them, it is does not have a fixed memory allocation or runtime per timestep. These numbers can change depending on the size and connectivity of the dynamic graph - which in turn depends on the number of objects in the scene and the associations made in the recent past. To keep the memory requirements to a minimum, we make use of sparse operations wherever possible - especially in our vertex update functions. To understand the memory requirements of the model described in Section~\ref{sec:state-of-art}, we plot histograms of allocated GPU memory during training and inference (Figure~\ref{fig:train-mem-alloc}, \ref{fig:val-mem-alloc}). From these plots, we notice that despite some spread, the allocated memory remains well within reasonable limits during training and inference (550-900MB). This makes it possible to train and test our model on most modern desktop and laptop GPUs. 

Similarly, we plot a histogram of our model runtime on the KITTI MOT test set (Figure~\ref{fig:infer-runtime}) using an NVIDIA Titan X Maxwell GPU. This plot demonstrates that our entire framework is capable of operating in real-time, taking only 0.01 seconds on average to process an entire timestep. Finally, to examine the effects of the size of the dynamic graph on runtime, we plot the number of detections nodes in the graph versus the observed runtime in Figure~\ref{fig:objects-vs-runtime}. We can clearly see a monotonically increasing relationship between the two; but the runtime increases at a faster rate as more detection nodes are added. This does not pose a problem for the KITTI dataset, but for larger datasets with highly cluttered scenes, occasional pruning operations can be used to keep the memory allocation within bounds.

\section{Concluding Remarks}
This study proposes a tracking framework based on undirected graphs that represent the data association problem over multiple timesteps in the tracking-by-detection paradigm. In this framework, both individual detections and associations between pairs of them are represented as nodes in the graph. Furthermore, the graph is dynamic, where only detections (and their associations) within a certain temporal range are processed in a rolling window basis. This form of data representation offers any multi-object tracking model the following benefits -  multiple competing associations can be considered while scoring any given association from two detections, including associations  over  multiple  timesteps.  Information  can  be  stored and  propagated  across  many  timesteps  through  message passing operations. Mistakes in the past can be corrected as long as they are still part of the dynamic graph. 
False positives, duplicate and missed detections can be handled intrinsically as part of the model. 

To illustrate these benefits, we also present a message passing neural network (TrackMPNN) that operates on these dynamic undirected graphs, and train it on the KITTI dataset. Our proposed training and inference schemes make this possible with limited memory and computational bandwidth, and enable real-time operation despite large number of objects in the scene.
Experiments, qualitative examples and competitive results on popular MOT benchmarks for autonomous driving demonstrate the promise and uniqueness of the proposed approach. 

\section{Acknowledgements}
We are grateful to the Laboratory for Intelligent \& Safe Automobiles at UC San Diego for providing us with the resources and compute to run the experiments presented in the paper.


{\small
\bibliographystyle{ieee_fullname}
\bibliography{main}
}

\end{document}